\pdfoutput=1
\documentclass[letterpaper]{article}

\def\ARXIVAUTHORS{Xuanhua Yin, Shunqi Mao, Wei Guo, Chuanzhi Xu, Weidong Cai\corresponding}
\def\ARXIVAFFILIATIONS{%
School of Computer Science, The University of Sydney\\
\texttt{\{xuanhua.yin, shunqi.mao, wei.guo, chuanzhi.xu, tom.cai\}@sydney.edu.au}}

\def\ARXIVROOT{1}
\def\ARXIVVERSION{1}
\ifdefined\SUPPLEMENTROOT
\else
\ifdefined\ARXIVROOT
\else
\documentclass[letterpaper]{article} 
\fi
\fi
\ifdefined\ARXIVVERSION
\usepackage[preprint]{aaai2027}
\else
\usepackage[submission]{aaai2027}  
\fi
\usepackage[hyphens]{url}  
\usepackage{graphicx} 
\usepackage{natbib}  
\usepackage{caption} 
\usepackage{algorithm}
\usepackage{algorithmic}
\usepackage{booktabs}
\usepackage{colortbl}
\usepackage{tabularx}
\usepackage{amsmath}
\usepackage{amssymb}

\ifdefined\SUPPLEMENT
\title{Supplementary Material for\\Calibrate What You SHIP:\\Post-Selection Risk Control for Verifier-Guided Text-to-Image Generation}
\else
\title{Calibrate What You SHIP:\\Post-Selection Risk Control for Verifier-Guided Text-to-Image Generation}
\fi
\ifdefined\ARXIVVERSION
\author{\ARXIVAUTHORS}
\affiliations{\ARXIVAFFILIATIONS}
\else
\author{
    Anonymous Submission
}
\affiliations{}
\fi

\newif\ifshipmain
\newif\ifshipsupp
\ifdefined\SUPPLEMENT
  \shipmainfalse
  \shipsupptrue
\else
  \shipmaintrue
  \shipsuppfalse
\fi
\ifdefined\ARXIVVERSION
  \shipmaintrue
  \shipsupptrue
\fi

\begin{document}
\setcounter{secnumdepth}{2}

\maketitle
\ifshipmain

\begin{abstract}
Verifier-guided text-to-image systems increasingly use test-time search
to select, refine, or stop among multiple candidates, yet release
thresholds are often calibrated on individual images. This creates a
candidate-to-policy calibration mismatch: search changes both which
prompts receive an output and which candidate is released, so
candidate-level risk control need not imply control of released-output
risk. We formalize this estimand shift through prompt reweighting and
within-prompt selection, and introduce \textbf{SHIP},
\emph{Selection-aware Held-out calibration of Inference Policies}.
SHIP runs or replays the complete deployed policy on held-out prompts,
evaluates the image it actually releases using an independent target
judge, and selects the most permissive threshold whose risk upper bound
satisfies a prescribed budget. For replayable policies with a
prespecified threshold grid, simultaneous confidence control provides
finite-sample validity. Experiments across fixed, sequential, and
adaptive T2I inference procedures show that policy-level calibration
recovers lower-risk operating points while exposing policy-dependent
tradeoffs among risk, coverage, and compute. On GenEval2 with FLUX at
$N=16$, a pooled-candidate threshold yields released risk $0.310$,
whereas SHIP reduces it to $0.162$. Across 200 cached-stream splits, the
fixed-grid certificate has no target crossing. Reliable inference-time
scaling therefore requires calibrating the output distribution induced
by the complete deployed policy.
\end{abstract}

\section{Introduction}
Modern text-to-image (T2I) systems increasingly allocate test-time
compute to search over multiple generation paths rather than returning
a single independently sampled image. Depending on the inference
procedure, the system may explore different noise seeds, rank partially
denoised drafts, rewrite prompts, refine unmet requirements, edit
intermediate images, stop early, or abstain~\citep{ma2025inference,kim2025das,li2025reflectdit,zhuo2025reflectionflow,chen2025t2icopilot,guo2026probeselect,jiang2026raise,rawal2026flashbon,qu2026adecot}. Fine-grained
vision-language evaluators guide these decisions by assessing object
counts, attributes, spatial relations, and actions~\citep{hu2023tifa,cho2024dsg,lin2024vqascore,wiles2025gecko,kamath2025geneval2}. Consequently, the
image delivered to the user is jointly determined by generation,
verification, search, and stopping. Reliability should therefore be
evaluated on the outputs released by the complete inference procedure,
rather than on independently generated candidates.

\begin{figure}[t]
\centering
\includegraphics[width=\columnwidth]{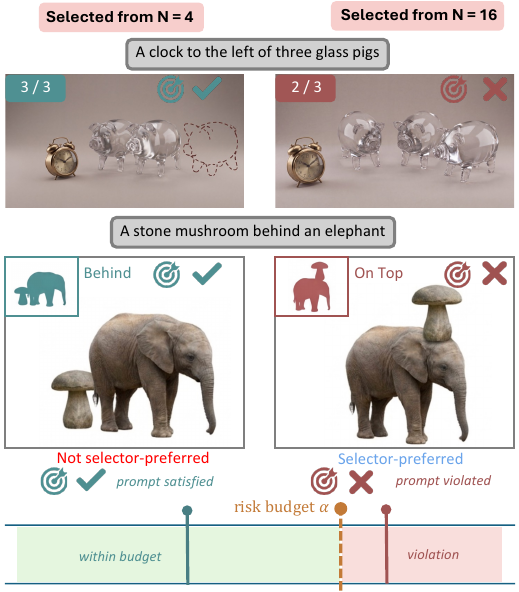}
\caption{Search can break candidate-level calibration. SHIP calibrates the output released by the complete policy.}
\label{fig:teaser}
\end{figure}

A common approach to improving reliability is to calibrate a verifier
threshold on held-out candidate images and release only candidates that
pass this threshold~\citep{angelopoulos2024crc}. Such calibration controls the expected loss of an
individual candidate conditioned on passing. In deployment, however,
the system does not release a generic passing candidate. It releases
the image returned by the complete search, selection, and stopping
procedure. Because this procedure changes the distribution of released
images, controlling candidate-level risk does not in general guarantee
control of the risk of the deployed inference policy. We refer to this
discrepancy as the \emph{candidate-to-policy calibration mismatch}.

Search creates this mismatch through two distinct mechanisms. First,
it changes which prompts receive an output. Under sequential stopping,
for example, a prompt is accepted if any attempt passes, so prompts
with different candidate-level pass probabilities have different
probabilities of appearing among the released outputs. Second, search
changes which image is released for each accepted prompt. Best-of-$N$
selection, first-hit stopping, tournaments, rewriting, and refinement
select or modify candidates according to an imperfect online verifier~\citep{gao2023overoptimization,coste2024ensembles},
so the released image need not follow the same conditional loss
distribution as a generic passing candidate. Search therefore changes
both the distribution of prompts represented among released outputs
and the conditional distribution of the image released for each prompt.

To this end, we introduce \textbf{SHIP},
\emph{Selection-aware Held-out calibration of Inference Policies}.
SHIP runs or replays the complete policy on held-out prompts, evaluates
the output it actually releases using an independent target judge, and
selects the most permissive threshold whose upper confidence bound
satisfies a prescribed risk budget. Threshold-independent candidate
streams support efficient replay, while history-dependent search is
evaluated through complete policy trajectories. For prespecified
threshold grids, simultaneous confidence bounds provide finite-sample
control for replayable policies~\citep{angelopoulos2021learn}.

Figure~\ref{fig:teaser} illustrates how verifier-guided search may
prefer an output that still violates the prompt, whereas SHIP calibrates
the final release decision under a target risk budget. We evaluate SHIP
across fixed, sequential, and adaptive T2I inference procedures,
including tournament selection, refinement, prompt rewriting,
trajectory selection, and image editing. The results recover feasible
released-output operating points and reveal policy-dependent tradeoffs
among risk, coverage, and inference cost. More broadly, search
determines the attainable performance frontier, while calibration
determines which points on that frontier can be released under a risk
constraint. Our contributions are summarized as follows:
\begin{itemize}
    \item We characterize a T2I-specific candidate-to-policy calibration
    mismatch, decompose it into prompt reweighting and within-prompt
    selection, and show that candidate-level risk control need not
    control released-output risk.

    \item We introduce \textbf{SHIP}, which calibrates complete release
    policies on held-out prompts and selects the most permissive
    threshold satisfying a prescribed released-output risk budget.

    \item We validate SHIP across fixed, sequential, and adaptive T2I
    search policies, multiple models and benchmarks, and human
    evaluation, revealing policy-dependent tradeoffs among risk,
    coverage, and inference cost.
\end{itemize}

\section{Related Work}
\paragraph{Text-to-Image Generation and Verification.}
Modern T2I systems span diffusion and flow-based generators, including SDXL, Stable Diffusion~3, and FLUX~\citep{rombach2022ldm,podell2024sdxl,esser2024sd3,lipman2023flowmatching,flux2024}. Their prompt following is increasingly assessed through fine-grained question answering, as in TIFA, DSG, VQAScore, Gecko, and GenEval2~\citep{hu2023tifa,cho2024dsg,lin2024vqascore,wiles2025gecko,kamath2025geneval2}, or through learned reward models such as ImageReward, PickScore, and CycleReward~\citep{xu2023imagereward,kirstain2023pickapic,bahng2025cyclereward}. These evaluators provide useful candidate-ranking signals, but do not by themselves characterize the loss of the image released after search. SHIP therefore treats the verifier as an interchangeable selector and evaluates policy outputs with a held-out target judge.

\paragraph{Verifier-Guided Inference-Time Scaling.}
Inference-time scaling searches seeds or trajectories, revises prompts, and edits images using verifier feedback~\citep{ma2025inference,kim2025das,li2025reflectdit,zhuo2025reflectionflow,chen2025t2icopilot}. Recent adaptive policies include RAISE refinement, ProbeSelect trajectory pruning, Flash-BoN draft tournaments, and ADE-CoT image editing~\citep{jiang2026raise,guo2026probeselect,rawal2026flashbon,qu2026adecot}. These methods improve how candidates are produced or selected. SHIP addresses the complementary reliability question of whether the output released by the resulting policy satisfies a target risk budget. We test this composition across fixed search, sequential stopping, refinement, rewriting, early trajectory selection, and editing. Because verifier bias and self-consistency can distort selection~\citep{zheng2023judging,wang2024fair}, our protocol separates the online selector from the offline target judge.

\paragraph{Selective Risk Control for Deployed Policies.}
Conformal prediction and its risk-controlling extensions calibrate decisions on exchangeable data~\citep{vovk2005algorithmic,bates2021distribution,angelopoulos2021learn,angelopoulos2024crc}, while selective classification trades coverage for conditional accuracy through abstention~\citep{chow1970optimum,elyaniv2010foundations,geifman2017selective}. Recent work extends these ideas to bounded selective risks and generated outputs~\citep{xu2025scrc,bai2026score,yu2026jointcert,kladny2025scopegen,wang2025safer}. TRON jointly controls sampling and filtering risks for multimodal response sets, but its output is a set rather than one image selected by a release policy~\citep{wang2025tron}. Most closely, BOKBO calibrates execute-or-abstain decisions for verifier-best $K$-sample vision-language-action policies~\citep{singh2026bokbo}. We therefore do not claim policy-level calibration or its confidence bounds as new. Our contribution is the T2I-specific estimand analysis, its prompt-reweighting and within-prompt decomposition, and evidence across first-hit release, best-of-$N$, and history-dependent policies with continuous released-output loss.

\section{Method}
\subsection{Overview and Setting}
Let $p\sim\mathcal{P}$ be a deployment prompt and let $\omega$ collect the internal randomness of generation and search. A thresholded release policy $\pi_\tau(p,\omega)$ executes its complete inference procedure and returns either an image or the abstention symbol $\bot$, where $\tau\in\mathcal{T}\subset[0,1]$ is a selector-risk threshold. Its trajectory may contain sampled images, partially denoised drafts, verifier feedback, rewritten prompts, or intermediate edits. Every completed image $Y$ has an online selector risk $s(Y,p)\in[0,1]$ and a held-out target loss $\ell(Y,p)\in[0,1]$, where lower values are better and the policy never observes $\ell$.

We write $A_\tau(p,\omega)=\mathbf{1}\{\pi_\tau(p,\omega)\neq\bot\}$, where $A_\tau$ indicates whether the policy releases an output. The goal is to control the target loss among released outputs at level $\alpha$ with confidence $1-\delta$, where $\alpha\in(0,1)$ is the risk budget and $\delta\in(0,1)$ is the calibration failure probability. A fixed candidate stream is the important special case $\omega=\mathbf{Y}=Y_{1:N}$ with $Y_k\sim G(\cdot\mid p)$, where $N$ is the maximum budget and $k\in[N]$ indexes candidates.

SHIP aligns the calibration unit with the release policy. Each exchangeable observation is a held-out prompt together with a complete policy rollout, rather than one candidate pooled across prompts. As illustrated in Figure~\ref{fig:pipeline}, SHIP evaluates the exact release rule and returns the most permissive threshold $\tau^\star$ whose policy-level risk bound satisfies $\alpha$. Threshold-independent streams support counterfactual replay of many thresholds from one cache. History-dependent procedures are instead evaluated from complete trajectories generated by the deployed adaptive policy. This distinction changes computational reuse, not the released-output estimand.

\begin{figure*}[t]
\centering
\includegraphics[width=0.97\textwidth]{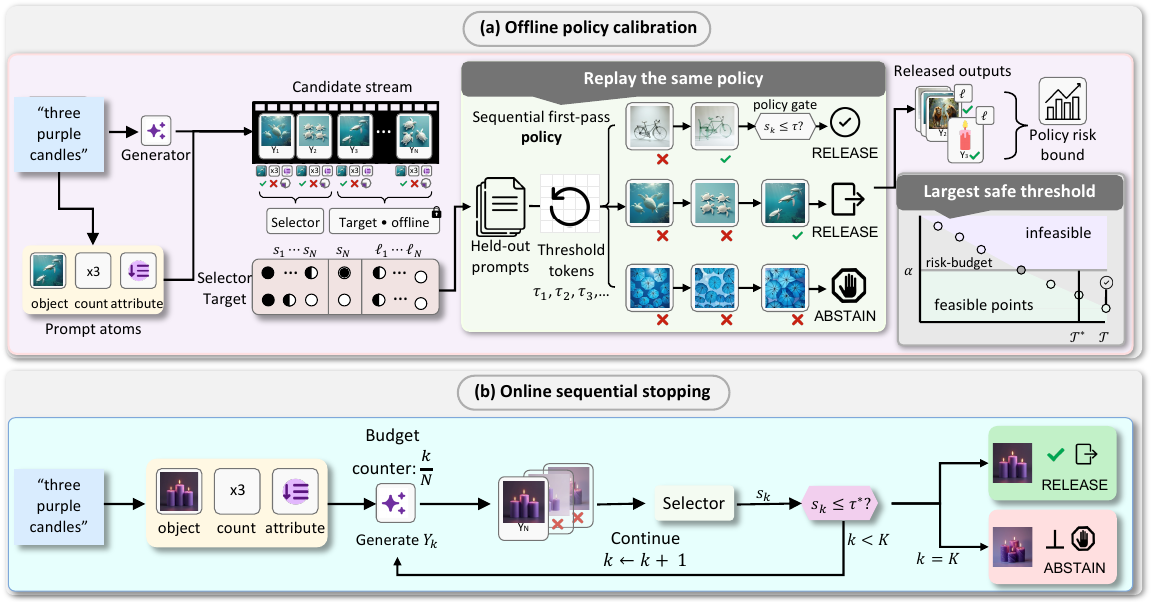}
\caption{SHIP calibration and deployment. \emph{a} Offline, the selector and held-out target score a candidate stream. SHIP replays the same sequential first-pass policy on held-out prompts across prespecified thresholds and selects the largest $\tau^\star$ whose released-output risk bound does not exceed $\alpha$. \emph{b} Online, sequential stopping releases the first candidate whose selector risk is at most $\tau^\star$, or abstains when the candidate budget is exhausted.}
\label{fig:pipeline}
\end{figure*}

\subsection{Atomic Prompt Verification}
To localize compositional failures, we represent each prompt by $\mathcal{A}(p)=\{a_j(p)\}_{j=1}^{m_p}$, where $a_j(p)$ is the $j$-th atomic requirement and $m_p=|\mathcal{A}(p)|$. The atoms cover existence, count, attribute binding, spatial relation, and action. Each atom becomes a yes/no question following TIFA and DSG~\citep{hu2023tifa,cho2024dsg}.

For a candidate $Y$, the continuous atom score is: 
\begin{equation}
q_j^{(v)}(Y,p)
=\Pr_{V_v}\!\left(\text{``Yes''}\mid Y,a_j(p)\right),
\qquad v\in\{\mathrm{sel},\mathrm{tgt}\},
\label{eq:atom-score}
\end{equation}
where $V_{\mathrm{sel}}$ is the selector VLM available at deployment, $V_{\mathrm{tgt}}$ is an independent target VLM used only offline, and $q_j^{(v)}(Y,p)\in[0,1]$ is the next-token probability that judge $V_v$ assigns to a positive answer. Continuous probabilities avoid the brittleness of hard yes/no decoding~\citep{lin2024vqascore}.

We aggregate atom scores using Soft-TIFA geometric-mean (Soft-TIFA-GM) risk:
\begin{equation}
\begin{aligned}
s(Y,p)
&=1-\left(\prod_{j=1}^{m_p}q_j^{(\mathrm{sel})}(Y,p)\right)^{1/m_p},\\
\ell(Y,p)
&=1-\left(\prod_{j=1}^{m_p}q_j^{(\mathrm{tgt})}(Y,p)\right)^{1/m_p},
\end{aligned}
\label{eq:risk}
\end{equation}
where $s(Y,p)$ is the online selector risk and $\ell(Y,p)$ is the held-out target loss. The geometric mean penalizes a low atom score multiplicatively, which is appropriate when one violated requirement can invalidate prompt following. We use Qwen2.5-VL~\citep{bai2025qwen25vl} for $V_{\mathrm{sel}}$ and Qwen3-VL~\citep{bai2025qwen3vl} for $V_{\mathrm{tgt}}$. InternVL3 provides an alternative-architecture robustness check~\citep{chen2024internvl,zhu2025internvl3}. Because the policy never observes $\ell$, calibration measures held-out risk rather than the selector's self-score.

\subsection{Candidate-to-Policy Calibration Mismatch}
Candidate-level calibration and deployment control different random variables:
\begin{equation}
\begin{aligned}
R_{\mathrm{img}}(\tau)
&=\mathbb{E}\!\left[\ell(Y,p)\mid s(Y,p)\le\tau\right],\\
R_\pi(\tau)
&=\mathbb{E}\!\left[
\ell\!\left(\pi_\tau(p,\omega),p\right)
\mid A_\tau(p,\omega)=1
\right],\\
C_\pi(\tau)
&=\Pr\!\left[A_\tau(p,\omega)=1\right],
\end{aligned}
\label{eq:estimand}
\end{equation}
where $Y\sim G(\cdot\mid p)$ is one candidate, $R_{\mathrm{img}}(\tau)$ is candidate risk conditional on passing, $R_\pi(\tau)$ is released-output risk, and $C_\pi(\tau)$ is coverage. Accept/reject evaluates only $Y_1$. Best-of-$N$ with rejection releases the minimum-risk candidate if it passes. Sequential stopping releases the first passing candidate. Adaptive policies may alter later candidates using earlier verifier feedback. The same threshold can therefore induce different released distributions across policies.

For a prompt $p$, let $q_\tau(p)=\Pr[s(Y,p)\le\tau\mid p]$ and $r_{\mathrm{img},\tau}(p)=\mathbb{E}[\ell(Y,p)\mid s(Y,p)\le\tau,p]$, where these terms are the candidate pass probability and conditional loss. Likewise, let $a_{\pi,\tau}(p)=\Pr[A_\tau(p,\omega)=1\mid p]$ and $r_{\pi,\tau}(p)=\mathbb{E}[\ell(\pi_\tau(p,\omega),p)\mid A_\tau(p,\omega)=1,p]$, where these terms are the policy acceptance probability and released-output loss.

\medskip\noindent\textbf{Proposition 1} (Candidate calibration is not policy calibration)\textbf{.}
\emph{Conditioning on the prompt gives}: 
\begin{equation}
\begin{aligned}
R_{\mathrm{img}}(\tau)
&=\frac{\mathbb{E}_p[q_\tau(p)r_{\mathrm{img},\tau}(p)]}
{\mathbb{E}_p[q_\tau(p)]},\\
R_\pi(\tau)
&=\frac{\mathbb{E}_p[a_{\pi,\tau}(p)r_{\pi,\tau}(p)]}
{\mathbb{E}_p[a_{\pi,\tau}(p)]},
\end{aligned}
\label{eq:reweighting}
\end{equation}
\emph{where $\mathbb{E}_p$ denotes expectation over deployment prompts. Consequently, $R_{\mathrm{img}}(\tau)\le\alpha$ does not imply $R_\pi(\tau)\le\alpha$. Selection changes the prompt weights from $q_\tau$ to $a_{\pi,\tau}$ and the conditional loss from $r_{\mathrm{img},\tau}$ to $r_{\pi,\tau}$. The latter term incorporates extreme-score selection, first-hit stopping, tournament outcomes, and history-dependent refinement.}

\subsection{Policy-Level Calibration and Validity}
SHIP evaluates the policy on $\mathcal{P}_{\mathrm{cal}}=\{p_i\}_{i=1}^{M}$, where $M$ is the number of held-out calibration prompts. For each $\tau\in\mathcal{T}$, it executes or exactly replays the policy with randomness $\omega_{i,\tau}$ and sets $A_i(\tau)=A_\tau(p_i,\omega_{i,\tau})$. When $A_i(\tau)=1$, it records $L_i(\tau)=\ell(\pi_\tau(p_i,\omega_{i,\tau}),p_i)$, where $L_i(\tau)\in[0,1]$ is the loss of the released output. Replayable policies may share a cached trajectory across thresholds. History-dependent policies use complete rollouts that preserve their native state transitions. Then: 
\begin{equation}
\begin{aligned}
n_\tau&=\sum_{i=1}^{M}A_i(\tau),\\
\widehat{R}_\pi(\tau)
&=\frac{1}{n_\tau}\sum_{i:A_i(\tau)=1}L_i(\tau),
\end{aligned}
\label{eq:empirical-risk}
\end{equation}
where $n_\tau$ is the accepted calibration count and $\widehat{R}_\pi(\tau)$ is the empirical policy risk for $n_\tau>0$.

SHIP computes: 
\begin{equation}
\begin{aligned}
U_{\delta'}(\tau)
&=\widehat{R}_\pi(\tau)
+\sqrt{\frac{\log(1/\delta')}{2n_\tau}},\\
\tau^\star
&=\max\{\tau\in\mathcal{T}:U_{\delta'}(\tau)\le\alpha\},
\end{aligned}
\label{eq:ship-calibration}
\end{equation}
where $\delta'\in(0,1)$ is the failure level assigned to one threshold, $U_{\delta'}(\tau)$ is the Hoeffding upper bound, and $\tau^\star$ is the most permissive feasible threshold. We set $U_{\delta'}(\tau)=+\infty$ for $n_\tau=0$ and return $\bot$ if none is feasible.

For finite-sample certification, SHIP fixes $\mathcal{T}$ before calibration and sets $\delta'=\delta/|\mathcal{T}|$, where $|\mathcal{T}|$ is the number of thresholds. This Bonferroni allocation makes all threshold bounds simultaneously valid and is a standard Learn-then-Test procedure~\citep{angelopoulos2021learn,bates2021distribution}. The confidence construction is not our methodological claim. Our contribution is to apply valid threshold selection to the losses produced by complete policy replay. Sharper selective-risk certificates~\citep{bai2026score,yu2026jointcert} can replace the Hoeffding bound without changing the policy-level estimand.

Certification is inexpensive when one threshold-independent stream can be replayed. If a threshold changes the state transitions of a history-dependent policy, exact evaluation may instead require a complete rollout for every prompt and threshold, giving naive offline cost proportional to $M|\mathcal{T}|$. We therefore report formal fixed-grid certification for replayable streams and treat the adaptive-policy operating points as descriptive. Reducing valid calibration cost for threshold-dependent search remains an open problem.

\medskip\noindent\textbf{Proposition 2} (Finite-sample validity)\textbf{.}
\emph{Assume the calibration prompts and their complete policy rollouts are independent and identically distributed draws from the deployment procedure, and let $\mathcal{T}$ be fixed independently of them. With $\delta'=\delta/|\mathcal{T}|$, any threshold $\tau^\star$ returned by~\eqref{eq:ship-calibration} satisfies}: 
\begin{equation}
\Pr\!\left[R_\pi(\tau^\star)\le\alpha\right]\ge 1-\delta,
\label{eq:validity}
\end{equation}
\emph{where the probability is over the calibration prompts and policy randomness.}

\noindent A proof is provided in the Supplementary Material.

\subsection{Sequential Calibrated Stopping}
Let $K_\tau(p,\mathbf{Y})=\min(\{k\in[N]:s(Y_k,p)\le\tau\}\cup\{+\infty\})$, where $K_\tau$ is the first passing index and equals $+\infty$ if none passes. The sequential policy is: 
\begin{equation}
\begin{aligned}
\pi_\tau^{\mathrm{seq}}(p,\mathbf{Y})
&=
\begin{cases}
Y_{K_\tau(p,\mathbf{Y})}, & K_\tau(p,\mathbf{Y})\le N,\\
\bot, & K_\tau(p,\mathbf{Y})>N,
\end{cases}\\
A_\tau^{\mathrm{seq}}(p,\mathbf{Y})
&=\mathbf{1}\!\left\{K_\tau(p,\mathbf{Y})\le N\right\}, 
\end{aligned}
\label{eq:sequential-policy}
\end{equation}
where $\pi_\tau^{\mathrm{seq}}$ releases the first passing candidate and $A_\tau^{\mathrm{seq}}$ indicates whether one exists within budget. Easy prompts stop early. Difficult prompts use up to $N$ attempts and otherwise abstain. Calibration evaluates this same first-hit rule, aligning the offline estimand with the online release.

Algorithm~\ref{alg:ship} summarizes calibration and deployment. For calibration prompt $p_i$, let $k_i(\tau)=K_\tau(p_i,\mathbf{Y}_i)$, where $k_i(\tau)$ is its first passing index at threshold $\tau$. The threshold grid is fixed before calibration so that the returned threshold inherits Proposition~2.

\begin{algorithm}[t]
\caption{SHIP calibration and sequential stopping}
\label{alg:ship}
\begin{algorithmic}[1]
\REQUIRE $\mathcal{P}_{\mathrm{cal}}=\{p_i\}_{i=1}^{M}$, generator $G$, selector $s$, target loss $\ell$
\REQUIRE budget $N$, target risk $\alpha$, failure level $\delta$, prespecified threshold grid $\mathcal{T}$
\STATE Sample $\mathbf{Y}_i=(Y_{i1},\ldots,Y_{iN})$ from $G(\cdot\mid p_i)$ for every $p_i$
\STATE Set $\delta'\gets\delta/|\mathcal{T}|$
\STATE $\tau^\star\gets\bot$
\FOR{$\tau\in\mathcal{T}$ in increasing order}
  \FOR{$i=1,\ldots,M$}
    \STATE $k_i(\tau)\gets K_\tau(p_i,\mathbf{Y}_i)$
    \IF{$k_i(\tau)\le N$}
      \STATE $A_i(\tau)\gets1$ and $L_i(\tau)\gets\ell(Y_{i,k_i(\tau)},p_i)$
    \ELSE
      \STATE $A_i(\tau)\gets0$
    \ENDIF
  \ENDFOR
  \STATE $n_\tau\gets\sum_{i=1}^{M}A_i(\tau)$
  \IF{$n_\tau>0$}
    \STATE Compute $\widehat{R}_\pi(\tau)$ by~\eqref{eq:empirical-risk} and $U_{\delta'}(\tau)$ by~\eqref{eq:ship-calibration}
    \IF{$U_{\delta'}(\tau)\le\alpha$}
      \STATE $\tau^\star\gets\tau$
    \ENDIF
  \ENDIF
\ENDFOR
\STATE \textbf{return} $\tau^\star$
\STATE \textbf{deploy:} abstain if $\tau^\star=\bot$. Otherwise release the first $Y_k$ with $s(Y_k,p)\le\tau^\star$ for $k\le N$, or abstain if none passes
\end{algorithmic}
\end{algorithm}

\section{Experiments and Results}
\subsection{Experimental Protocol}
\paragraph{Models and Benchmarks.}
GenEval2~\citep{kamath2025geneval2} provides the primary $800$-prompt evaluation. FLUX.1-dev~\citep{flux2024} supports the main fixed and adaptive comparisons. Stable Diffusion~3 Medium, SD3.5-Large~\citep{esser2024sd3}, and SDXL~\citep{podell2024sdxl} test backbone transfer. FLUX.1-Kontext-dev~\citep{labs2025fluxkontext} supports image editing. GenAI-Bench~\citep{li2024genaibench}, T2I-CompBench~\citep{huang2023t2icompbench}, DSG-1k~\citep{cho2024dsg}, and DPG-Bench~\citep{hu2024ella} test prompt-distribution transfer. By default, Qwen2.5-VL-7B-Instruct and Qwen3-VL-8B-Instruct provide selector risk and target loss, respectively. InternVL3-8B is the alternative-family target.

\paragraph{Protocol and Metrics.}
All primary comparisons use disjoint $400/400$ calibration and test prompts, target risk $\alpha{=}0.30$, and calibration failure probability $\delta{=}0.10$. The main tables report less conservative empirical operating points obtained with pointwise Hoeffding bounds. We call an operating point \emph{calibration-feasible} when its calibration UCB does not exceed $\alpha$. This empirical designation does not inherit Proposition~2 because pointwise bounds do not account for threshold search. The finite-sample certification study applies Algorithm~\ref{alg:ship} with the prespecified grid $\mathcal{T}=\{0,0.005,\ldots,1\}$ and $|\mathcal{T}|=201$. Coverage is the fraction of prompts receiving an output. Released-output risk is mean held-out target loss conditioned on release. Compute is reported as full-generation equivalents, total online VLM calls, and wall-clock time. Prompt bootstrap intervals preserve complete policy trajectories.

\paragraph{Comparisons.}
Fixed-search comparisons include random release, best-of-$N$ rejection, and sequential first-pass release. Adaptive policies include the Flash-BoN VLM tournament~\citep{rawal2026flashbon}, RAISE refinement~\citep{jiang2026raise}, history-dependent prompt rewriting, ProbeSelect early trajectory selection~\citep{guo2026probeselect}, and ADE-CoT image editing~\citep{qu2026adecot}. The pooled-candidate heuristic fits a threshold after treating all completed candidates as observations, although candidates from the same prompt are dependent. It is included as a common descriptive practice rather than a prompt-level certificate. A one-candidate-per-prompt sensitivity removes this pseudoreplication. SHIP instead fits the threshold to outputs of the complete policy under the same split, selector, and target judge. The Supplementary Document reports the prompt-unit sensitivity and compares our adaptive-policy reproductions with their original metrics.

\begin{figure*}[t]
\centering
\includegraphics[width=0.90\textwidth]{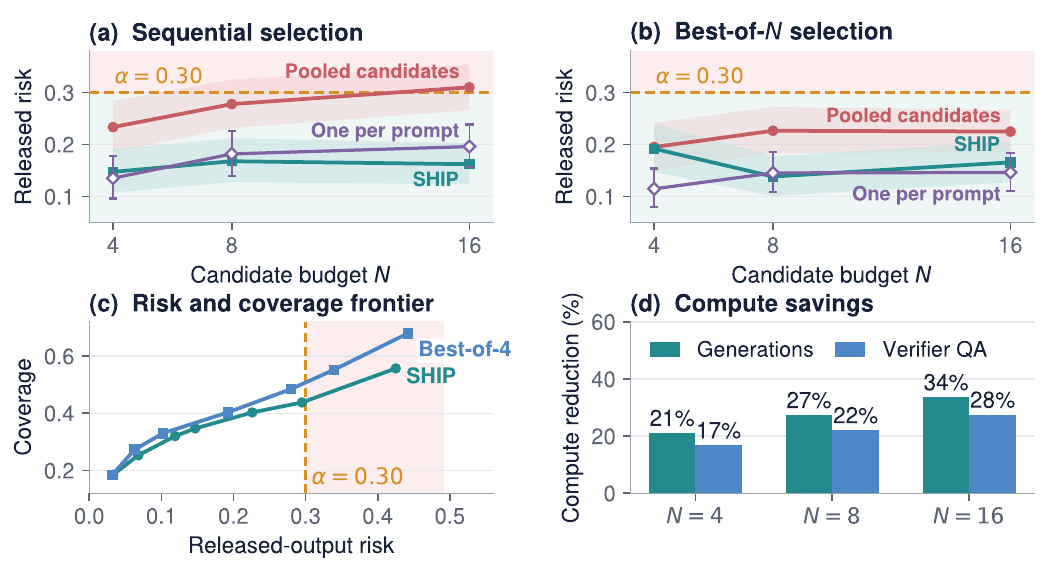}
\caption{Calibration depends on the statistical unit. \emph{a,b} Released risk with prompt-bootstrap intervals. Pooling ignores within-prompt dependence, whereas SHIP calibrates policy outputs. \emph{c} SHIP risk--coverage frontier. \emph{d} Sequential compute reduction relative to best-of-$N$ at matched risk.}
\label{fig:mismatch}
\end{figure*}

\begin{figure}[tb]
\centering
\begin{tabular}{@{}c@{\hspace{0.5em}}c@{}}
\multicolumn{2}{c}{\textbf{\small (a) Stop sign above bed}}\\[-0.15em]
\scriptsize Reject & \scriptsize First passing candidate\\[-0.15em]
\includegraphics[width=0.29\columnwidth,trim=0bp 26.5bp 195.3bp 19.7bp,clip]{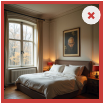} &
\includegraphics[width=0.29\columnwidth,trim=68.4bp 26.5bp 126.4bp 19.7bp,clip]{Figures/qualitative_filtering_examples.pdf}\\[0.3em]
\multicolumn{2}{c}{\textbf{\small (b) Chair beside chair}}\\[-0.15em]
\scriptsize Reject & \scriptsize First passing candidate\\[-0.15em]
\includegraphics[width=0.29\columnwidth,trim=127.0bp 26.5bp 68.3bp 19.7bp,clip]{Figures/qualitative_filtering_examples.pdf} &
\includegraphics[width=0.29\columnwidth,trim=195.2bp 26.5bp 0bp 19.7bp,clip]{Figures/qualitative_filtering_examples.pdf}
\end{tabular}
\caption{Four-image filtering examples at $\tau{=}0.30$. SHIP rejects the left candidates and releases the first passing candidates on the right.}
\label{fig:qualitative-filtering}
\end{figure}

\begin{table*}[t]
\centering
\small
\setlength{\tabcolsep}{2.4pt}
\begin{tabularx}{\textwidth}{@{}>{\raggedright\arraybackslash}Xlrrrrrr@{}}
\toprule
Search policy & Cal. unit & UCB$\downarrow$ & Cov.$\uparrow$ & Risk [95\% CI]$\downarrow$ & Gen.$\downarrow$ & VLM$\downarrow$ & Sec.$\downarrow$ \\
\midrule
\multicolumn{8}{@{}l}{\emph{Naive pooled-candidate threshold, within-prompt dependence ignored}} \\
\cmidrule(lr){1-8}
Sequential first-pass & Pooled cand. & 0.420 & 0.573 & 0.310 [0.267, 0.355] & 9.10 & 75.3 & 66.7 \\
Flash-BoN tournament & Pooled cand. & 0.472 & 0.598 & 0.361 [0.317, 0.405] & 4.52 & 133.8 & 45.2 \\
RAISE refinement & Pooled cand. & 0.355 & 0.688 & 0.244 [0.207, 0.281] & 18.60 & 146.8 & 142.3 \\
History-dependent rewriting & Pooled cand. & 0.439 & 0.535 & 0.328 [0.282, 0.374] & 5.40 & 46.0 & 49.4 \\
\midrule
\multicolumn{8}{@{}l}{\emph{Threshold calibrated on the complete deployed policy}} \\
\cmidrule(lr){1-8}
Sequential SHIP & Policy & 0.299 & 0.438 & \textbf{0.162 [0.125, 0.203]} & 10.68 & 87.6 & 78.2 \\
Flash-BoN + SHIP & Policy & 0.298 & 0.462 & 0.167 [0.130, 0.204] & \textbf{4.52} & 133.8 & \textbf{45.2} \\
RAISE + SHIP & Policy & \textbf{0.297} & \textbf{0.612} & 0.208 [0.172, 0.244] & 18.60 & 146.8 & 142.3 \\
Rewriting + SHIP & Policy & 0.299 & 0.401 & 0.176 [0.134, 0.218] & 5.40 & \textbf{46.0} & 49.4 \\
\bottomrule
\end{tabularx}
\caption{Policy-level calibration across adaptive FLUX search on the shared GenEval2 split ($\alpha{=}0.30$, $\delta{=}0.10$). Intervals use prompt bootstrap. UCBs are descriptive. Pooled rows also ignore within-prompt dependence and are not certificates. VLM calls include search verification.}
\label{tab:adaptive-main}
\end{table*}

\begin{figure}[t]
\centering
\includegraphics[width=0.82\columnwidth]{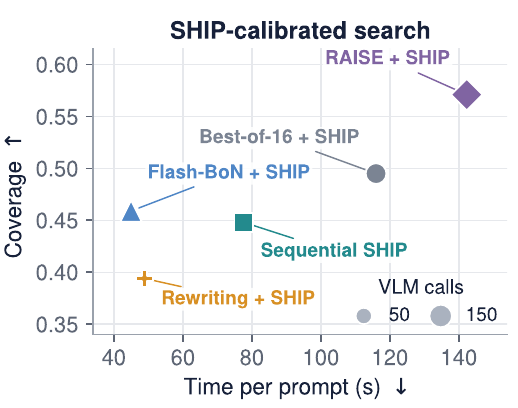}
\caption{Calibration-selected SHIP tradeoffs at test risk $0.166$. Higher is better, leftward is faster, and marker area denotes VLM calls. RAISE maximizes coverage, while Flash-BoN minimizes time. Thresholds are fit without test outputs.}
\label{fig:adaptive-frontier}
\end{figure}

\begin{table}[t]
\centering
\small
\setlength{\tabcolsep}{3.2pt}
\begin{tabularx}{\columnwidth}{@{}>{\raggedright\arraybackslash}Xrrr@{}}
\toprule
Policy & Cov.$\uparrow$ & Atomic loss$\downarrow$ & Any-failure$\downarrow$ \\
\midrule
Pooled-candidate & 0.573 & 0.112 & 0.380 \\
\textbf{Sequential SHIP} & 0.438 & \textbf{0.076} & \textbf{0.280} \\
$\Delta$ & $-0.135$ & $-0.036$ & $-0.100$ \\
\bottomrule
\end{tabularx}
\caption{Human audit of released outputs. $\Delta$ is SHIP minus the pooled-candidate baseline.}
\label{tab:human-main}
\end{table}

\subsection{Candidate-to-Policy Mismatch across Search Policies}
\paragraph{Sequential Search Exposes the Estimand Shift.}
Figure~\ref{fig:mismatch}a isolates the effect on a fixed FLUX candidate stream. The naive pooled threshold gives sequential released risks of $0.234$, $0.278$, and $0.310$ as $N$ grows from $4$ to $16$. At $N{=}16$, the interval $[0.267,0.355]$ overlaps the target, so the crossing is not statistically resolved. Pooling also confounds the estimand shift with within-prompt dependence: one-candidate-per-prompt calibration gives risk $0.196$, versus $0.162$ for policy calibration. Thus the magnitude is design-sensitive, but the candidate and policy objectives remain distinct. The Supplementary Document gives prompt-unit and independent-stream analyses plus a controlled example approaching gap $1-\alpha$.

\paragraph{The Mismatch Generalizes beyond Fixed Search.}
Table~\ref{tab:adaptive-main} applies the same descriptive protocol to four FLUX policies. Naive pooled thresholds produce pointwise UCBs from $0.355$ to $0.472$. Sequential first-pass release, Flash-BoN, and history-dependent rewriting have test risks of $0.310$, $0.361$, and $0.328$, respectively. RAISE has a lower test point estimate of $0.244$, although its calibration UCB is $0.355$. These rows diagnose how a pooled threshold behaves after deployment search. They do not constitute a valid prompt-level candidate certificate.

SHIP reduces every descriptive policy-level UCB to at most $0.299$, with test risk at most $0.208$. Search design sets the attainable coverage and cost profile, while policy-level calibration determines which outputs may be released, as illustrated in Figure~\ref{fig:qualitative-filtering}.

\subsection{Risk, Coverage, and Compute Tradeoffs}
\paragraph{Coverage Is a Tunable Operating Characteristic.}
Figure~\ref{fig:mismatch}c sweeps $\alpha$ rather than selecting a favorable threshold. Coverage rises from $0.253$ to $0.557$ as $\alpha$ increases from $0.20$ to $0.50$, while risk tracks the requested budget. Best-of-$4$ rejection covers more prompts but always uses four generations. Lower coverage at a strict budget therefore reflects deliberate abstention.

\paragraph{Search Policy Determines the Efficiency Profile.}
At risk $0.166$, Figure~\ref{fig:adaptive-frontier} shows that Flash-BoN is fastest, RAISE has the highest coverage, and rewriting uses the fewest VLM calls. Sequential SHIP is intermediate, and no policy is best under every cost measure.

\subsection{Robustness and Generalization}
\paragraph{Adaptive Policies Transfer across Backbones and Tasks.}
On SD3.5-Large, ProbeSelect with SHIP reaches risk $0.196$, UCB $0.298$, and coverage $0.428$ using $1.80$ rather than $5.00$ full-denoise equivalents. On FLUX.1-Kontext-dev, ADE-CoT with SHIP reaches edit loss $0.171$ and UCB $0.298$, while candidate calibration is infeasible. These results support calibration transfer, not lower native editing cost. Complete operating points appear in the Supplementary Document.

\paragraph{Additional Robustness Checks.}
Sequential SHIP remains below target across backbones, but SDXL coverage falls to $0.113$ and SD3 transfer requires recalibration at risk $0.303$. At $N{=}16$, the certificate shows no target crossing across $200$ cached-stream splits with mean risk $0.148$ and coverage $0.391$. These splits reuse generations and omit rerun variability. Full results are supplementary.

\paragraph{Human Released-Output Audit.}
Human atomic loss falls from $0.112$ to $0.076$ and strict any-failure from $0.380$ to $0.280$ (Table~\ref{tab:human-main}). Because calibration uses machine loss, this corroborates improvement but is not a human-risk certificate. The Supplementary Document reports the audit protocol, aggregation definitions, and aggregate agreement statistics.

\subsection{Evaluator and Failure Analysis}
\paragraph{Why Fine-Grained Risk Matters.}
Soft-TIFA-GM gives Spearman $0.928$, AUROC $0.975$, and the lowest non-oracle risk. Scalar rewards yield no feasible set, so ranking alone is insufficient.

\paragraph{Residual Failures.}
Failures concentrate in counting, spatial, and action. Prompts with $9$ to $10$ atoms reach risk $0.398$ at $0.05$ coverage, showing that average control does not ensure subgroup service.

\section{Conclusion}
Verifier-guided generation releases a policy-selected output rather than an independent candidate. SHIP calibrates this deployed object by replaying the policy on held-out prompts and bounding released-image loss. Across fixed and adaptive search, policy-level calibration better characterized released-output risk than pooled-candidate thresholds and recovered feasible risk, coverage, and compute tradeoffs. Search determines the attainable quality, coverage, and cost frontier, while calibration determines which operating points may be released.

Validity requires a prespecified grid, exchangeability, bounded loss, and faithful replay. Threshold-dependent adaptive search remains descriptive because exact calibration may require a rollout per prompt and threshold. A held-out VLM defines loss, making the human audit corroborative rather than a human-risk certificate. SHIP controls average rather than subgroup risk. Evaluator drift, hard-prompt abstention, and distribution shift remain boundaries.

Future work should develop sharper bounds, group-aware objectives, and shift detection. More broadly, reliability should be calibrated on the complete deployed decision process that releases an output, not on upstream candidates.

\clearpage
{\small
\bibliography{aaai2027}
}
\fi

\ifshipsupp
\setcounter{section}{0}
\renewcommand{\thesection}{\Alph{section}}
\ifdefined\ARXIVVERSION
\clearpage
\twocolumn[
\begin{center}
{\LARGE\bfseries Supplementary Material\par}
\vspace{1.0em}
\end{center}
]
\fi
\section*{Overview}
This document expands the evidence behind the main paper without changing its
evaluation protocol or reported operating points. Section~\ref{sec:supp-protocol} specifies the
generation, verification, calibration, and uncertainty procedures.
Section~\ref{sec:supp-formal} states the formal scope of SHIP and proves Proposition~2.
Section~\ref{sec:supp-mismatch} diagnoses the candidate-to-policy mismatch through mechanism,
calibration-unit and independent-stream analyses. Section~\ref{sec:supp-adaptive}
gives the adaptive-policy results and Section~\ref{sec:supp-robustness} tests robustness.
Sections~\ref{sec:supp-qualitative} to~\ref{sec:supp-evaluator} provide qualitative, fixed-search, human, and evaluator
evidence. Section~\ref{sec:supp-boundaries} states the remaining deployment boundaries. Unless
stated otherwise, experiments use GenEval2
with FLUX.1-dev, Qwen2.5-VL Soft-TIFA-GM selector risk, Qwen3-VL Soft-TIFA-GM
target loss, $\alpha{=}0.30$, $\delta{=}0.10$, and disjoint sets of $400$
calibration prompts and $400$ test prompts.

\section{Evaluation Protocol and Implementation Details}
\label{sec:supp-protocol}

\paragraph{Generation and verification.}
FLUX.1-dev uses $28$ denoising steps, guidance $3.5$, and
$1024\times1024$ resolution. SD3 Medium uses $28$ steps and classifier-free
guidance $7.0$. SDXL uses $30$ steps and guidance $7.0$ at the same
resolution. Candidate index $k$ corresponds to generation seed $k-1$.
The selector checkpoint is Qwen2.5-VL-7B-Instruct and the primary target is
Qwen3-VL-8B-Instruct. InternVL3-8B supplies the alternative-family target.
Atom questions and expected answers are frozen in the GenEval2 metadata.
Each query appends ``Answer in one word.'' The evaluator sums the next-token
probabilities of valid capitalization, whitespace, and numeric variants before
geometric-mean aggregation.

\paragraph{Calibration units and statistical summaries.}
One prompt and its complete policy trajectory form one exchangeable unit for
SHIP. The pooled-candidate heuristic instead treats completed images as
observations even when several images share a prompt. We label this heuristic
naive and empirical throughout. The one-candidate-per-prompt sensitivity
restores the prompt as the sampling unit, but its pointwise threshold scan is
not a simultaneous certificate. Test intervals use prompt bootstrap resampling
of complete release or abstention outcomes. Human-policy differences use the
same prompt-level paired resampling. Zero coverage denotes an infeasible
operating point and is never interpreted as zero deployment risk.
Unless a table states otherwise, each fixed-stream operating point is computed
once on the frozen candidate stream and each interval uses $5000$ prompt-level
bootstrap replicates with a fixed analysis seed. Candidate-stream replications
and cached-stream split studies state their seed ranges and repetition counts
separately. Bootstrap resampling quantifies prompt-sampling uncertainty and is
not counted as an independent generation run.

\paragraph{Hardware.}
Core generation and scoring runs use one NVIDIA RTX 5090 with 32 GB memory,
driver 580.159.03, CUDA 12.8, and PyTorch 2.10.0. Baseline-specific
environments use the package versions required by their official
implementations. Reported seconds are single-GPU wall-clock measurements.

\paragraph{Adaptive-Policy Implementations.}
Flash-BoN uses its full VLM tournament over inexpensive drafts. RAISE uses its
analyzer, rewriter, verifier, generation, and editing actions. The
history-dependent rewriting ablation retains failure-conditioned prompt
rewriting and generation while disabling image editing. ProbeSelect launches
five SD3.5-Large trajectories and retains one after the probe at $20\%$ of
denoising. It uses $28$ steps and classifier-free guidance $7.0$. ADE-CoT
follows its official $N{=}32$ editing protocol. Every completed candidate and
released output in the common comparison is rescored by the same selector and
target judge. Reported VLM calls include search-native verification and the
release gate.

\FloatBarrier
\section{Formal Scope and Finite-Sample Validity}
\label{sec:supp-formal}

\paragraph{Proof of Proposition~2.}
For any fixed $\tau$ and $n_\tau>0$, the accepted losses are bounded in $[0,1]$. Hoeffding's inequality therefore gives $\Pr[R_\pi(\tau)>U_{\delta'}(\tau)]\le\delta'$. A union bound over the $|\mathcal{T}|$ prespecified thresholds makes every bound valid simultaneously with probability at least $1-\delta$. On this event, the selected threshold obeys $R_\pi(\tau^\star)\le U_{\delta'}(\tau^\star)\le\alpha$. $\square$

\begin{table*}[t]
\centering
\small
\setlength{\tabcolsep}{3.7pt}
\renewcommand{\arraystretch}{1.08}
\begin{tabular}{@{}p{1.45cm}p{2.35cm}p{2.65cm}p{2.65cm}p{2.65cm}@{}}
\toprule
Method & Calibration unit & Released object and loss & Policy family & Statistical statement \\
\midrule
Learn-then-Test & Exchangeable examples for a parameterized algorithm & User-defined bounded risk & Any prespecified algorithm family & Multiple-testing risk control over parameters \\
SCOPE-Gen & Generated candidate sets & At least one admissible set member & Sequential generation and greedy set filters & Conformal admissibility control \\
SAFER & Open-ended language prompts & Correct-answer inclusion in a filtered set & Abstention-aware sampling and filtering & Separate sampling and filtering risk controls \\
BOKBO & VLA instruction with $K$ actions & Unsafe execution among non-abstained decisions & Verifier-best action and abstention & Finite-sample decision-level risk bound \\
SHIP & Prompt with complete policy rollout & Continuous loss of one released image & Fixed or history-dependent search with abstention & Learn-then-Test applied to policy-output loss \\
\bottomrule
\end{tabular}
\caption{Formal scope comparison. SHIP does not introduce a new concentration inequality or the general idea of decision-level calibration. Its contribution is the candidate-to-policy estimand analysis and a T2I instantiation that covers fixed and adaptive release policies.}
\label{tab:scope}
\end{table*}

Table~\ref{tab:scope} separates the statistical tool from the paper's
application-level contribution. Learn-then-Test supplies simultaneous
parameter selection. SHIP defines the calibrated observation as the output of
the complete T2I release policy. SCOPE-Gen and SAFER control set-valued or
filtered generation objects. BOKBO controls an execute-or-abstain decision for
verifier-best actions. These objects are related, but none makes a
candidate-level threshold automatically valid after a different search and
stopping policy is composed with it.

\FloatBarrier
\section{Diagnosing the Candidate-to-Policy Mismatch}
\label{sec:supp-mismatch}

\paragraph{Mechanism decomposition.}
Table~\ref{tab:decomposition} implements the decomposition in Proposition~1.
For sequential release at $N{=}16$, changing candidate pass weights to policy
acceptance weights contributes $+0.108$ to risk. Prompt bootstrap gives an
interval from $0.083$ to $0.133$. Replacing the mean passing-candidate loss by
the first released candidate contributes $-0.009$, with an interval from
$-0.035$ to $0.017$. The observed sequential mismatch is therefore dominated
by prompt reweighting in this experiment. Best-of-$N$ selection has a more
negative within-prompt component, which offsets part of the same reweighting.

\begin{table*}[t]
\centering
\small
\setlength{\tabcolsep}{6pt}
\renewcommand{\arraystretch}{1.08}
\begin{tabular}{@{}clrrrrrr@{}}
\toprule
$N$ & Policy & Candidate & Reweighted & Released & Prompt weights & Within prompt & Total gap \\
\midrule
4 & Sequential & 0.195 & 0.246 & 0.234 & +0.051 & -0.012 & +0.038 \\
8 & Sequential & 0.204 & 0.295 & 0.278 & +0.091 & -0.018 & +0.073 \\
16 & Sequential & 0.211 & 0.319 & 0.310 & +0.108 & -0.009 & +0.099 \\
4 & Best-of-$N$ & 0.195 & 0.246 & 0.195 & +0.051 & -0.051 & 0.000 \\
8 & Best-of-$N$ & 0.204 & 0.295 & 0.227 & +0.091 & -0.069 & +0.022 \\
16 & Best-of-$N$ & 0.211 & 0.319 & 0.225 & +0.108 & -0.094 & +0.014 \\
\bottomrule
\end{tabular}
\caption{Mechanism decomposition at the pooled candidate threshold. Candidate risk uses candidate pass weights. Reweighted risk substitutes policy acceptance weights while retaining each prompt's mean passing-candidate loss. Released risk then substitutes the policy-selected loss. Prompt bootstrap intervals for the main $N{=}16$ sequential components are reported in the text.}
\label{tab:decomposition}
\end{table*}

\paragraph{Sensitivity to the candidate calibration unit.}
The pooled heuristic combines $16$ correlated candidates from each prompt and
thus reports $6400$ calibration observations. Table~\ref{tab:prompt-unit}
compares it with two prompt-level alternatives and direct policy replay on the
same fixed $N{=}16$ stream. The pooled sequential row reproduces the risk
$0.310$ in the main paper. Choosing one fixed candidate per prompt reduces the
point estimate to $0.196$, so the numerical budget crossing depends on the
candidate calibration design. Direct policy replay targets the deployed
first-pass output and gives risk $0.162$. The comparison supports an estimand
distinction, not a claim that every candidate-level construction must cross
$\alpha$ on every finite dataset.

\begin{table*}[t]
\centering
\small
\setlength{\tabcolsep}{4.2pt}
\renewcommand{\arraystretch}{1.08}
\begin{tabular}{@{}llrrrrr@{}}
\toprule
Policy & Calibration design & Cal. units & $\tau$ & Cov.$\uparrow$ & Risk and $95\%$ CI$\downarrow$ & Cal. UCB$\downarrow$ \\
\midrule
Sequential & Naive pooled candidates & 6400 & 0.450 & 0.573 & 0.310 [0.267, 0.355] & 0.300 \\
Sequential & One fixed candidate per prompt & 400 & 0.322 & 0.473 & 0.196 [0.156, 0.239] & 0.292 \\
Sequential & One hash-selected candidate per prompt & 400 & 0.268 & 0.453 & 0.183 [0.143, 0.225] & 0.297 \\
\textbf{Sequential SHIP} & \textbf{Complete policy replay} & \textbf{400} & \textbf{0.244} & \textbf{0.438} & \textbf{0.162 [0.125, 0.203]} & \textbf{0.299} \\
\midrule
Best-of-$16$ & Naive pooled candidates & 6400 & 0.450 & 0.573 & 0.225 [0.186, 0.267] & 0.300 \\
Best-of-$16$ + SHIP & Complete policy replay & 400 & 0.348 & 0.495 & 0.166 [0.127, 0.206] & 0.299 \\
\bottomrule
\end{tabular}
\caption{Calibration-unit sensitivity on the fixed FLUX $N{=}16$ stream. All thresholds use a pointwise Hoeffding scan on the same $400$ calibration prompts and are evaluated on the same $400$ test prompts. The pooled heuristic ignores within-prompt dependence. These empirical operating points do not inherit the simultaneous guarantee of Proposition~2.}
\label{tab:prompt-unit}
\end{table*}

\paragraph{Independent candidate-stream replication.}
To test dependence on generation randomness, we repeat the $N{=}16$
comparison on two nonoverlapping seed streams over the same frozen subset of
$96$ calibration and $104$ test prompts. Table~\ref{tab:independent-stream}
shows the same ordering in both streams. The smaller subset produces wider
intervals, so this is replication evidence rather than a replacement for the
full $800$-prompt experiment.

\paragraph{Representative release decisions.}
Figure~\ref{fig:policy-filtering-cases} shows two real prompt-level candidate
streams with different relation failures. In each case, the policy rejects an
early high-risk image and releases a later candidate that satisfies the stated
relation. The examples visualize the released object evaluated by SHIP rather
than serving as additional quantitative evidence.

\begin{figure}[H]
\centering
\includegraphics[width=0.98\columnwidth]{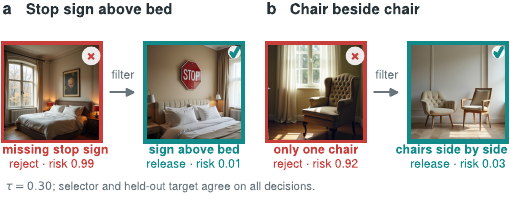}
\caption{Real FLUX release decisions. \emph{a} The first image omits the stop
sign, while the released image places it above the bed. \emph{b} The first
image contains only one chair, while the released image contains two chairs
side by side. Borders and labels encode the frozen selector decision. Image
content is shown without retouching or regeneration.}
\label{fig:policy-filtering-cases}
\end{figure}

\begin{table}[t]
\centering
\small
\setlength{\tabcolsep}{3.1pt}
\renewcommand{\arraystretch}{1.06}
\begin{tabular}{@{}lrrrr@{}}
\toprule
Seeds & Pooled cov. & Pooled risk & SHIP cov. & SHIP risk \\
\midrule
$0$ to $15$ & 0.587 & 0.340 & 0.413 & 0.182 \\
$16$ to $31$ & 0.510 & 0.299 & 0.327 & 0.103 \\
\bottomrule
\end{tabular}
\caption{Independent $N{=}16$ candidate-stream replication. The paired risk differences are $-0.158$ with interval $[-0.242,-0.079]$ for seeds $0$ to $15$, and $-0.196$ with interval $[-0.298,-0.097]$ for seeds $16$ to $31$.}
\label{tab:independent-stream}
\end{table}

\FloatBarrier
\section{Adaptive Search and Editing Policies}
\label{sec:supp-adaptive}

\begin{table*}[t]
\centering
\small
\setlength{\tabcolsep}{5.2pt}
\renewcommand{\arraystretch}{1.08}
\begin{tabular}{@{}lrrrrrr@{}}
\toprule
Policy with SHIP & Risk$\downarrow$ & Cov.$\uparrow$ & Gen-eq.$\downarrow$ & VLM/p$\downarrow$ & Sec./p$\downarrow$ & Primary profile \\
\midrule
Best-of-$16$ & 0.166 & 0.495 & 16.00 & 120.3 & 116.0 & Reference \\
Sequential & 0.166 & 0.448 & 10.61 & 87.1 & 77.7 & Balanced \\
Flash-BoN & 0.166 & 0.459 & \textbf{4.52} & 133.8 & \textbf{45.0} & Fastest \\
RAISE & 0.166 & \textbf{0.571} & 18.60 & 146.8 & 142.3 & Highest coverage \\
Rewriting & 0.166 & 0.394 & 5.40 & \textbf{46.0} & 48.9 & Fewest VLM calls \\
\bottomrule
\end{tabular}
\caption{Matched-risk efficiency across FLUX policies. Operating points are selected on the calibration split to match best-of-$16$. Their test risks round to $0.166$. The table reports complementary costs without assigning one Pareto label across incompatible resource axes.}
\label{tab:matched-risk}
\end{table*}

\begin{table*}[t]
\centering
\small
\setlength{\tabcolsep}{3.4pt}
\renewcommand{\arraystretch}{1.08}
\begin{tabular}{@{}llrrrrrr@{}}
\toprule
Search policy & Calibration unit & Cal. UCB$\downarrow$ & Cov.$\uparrow$ & Risk and $95\%$ CI$\downarrow$ & Denoise-eq.$\downarrow$ & QA/p$\downarrow$ & Sec./p$\downarrow$ \\
\midrule
Full-denoise best-of-$5$ & Candidate & 0.394 & 0.512 & 0.283 [0.236, 0.331] & 5.00 & 37.6 & 41.3 \\
ProbeSelect $5{\rightarrow}1$ & None & N/A & 1.000 & 0.377 [0.339, 0.416] & 1.80 & 0.0 & 13.8 \\
ProbeSelect $5{\rightarrow}1$ & Candidate & 0.429 & 0.598 & 0.312 [0.267, 0.358] & 1.80 & 7.5 & 14.6 \\
Full-denoise best-of-$5$ + SHIP & Policy & \textbf{0.298} & \textbf{0.445} & 0.191 [0.148, 0.235] & 5.00 & 37.6 & 41.3 \\
Sequential SHIP & Policy & 0.299 & 0.406 & \textbf{0.184 [0.139, 0.230]} & 3.42 & 25.7 & 28.3 \\
\textbf{ProbeSelect $5{\rightarrow}1$ + SHIP} & Policy & \textbf{0.298} & 0.428 & 0.196 [0.152, 0.241] & \textbf{1.80} & \textbf{7.5} & \textbf{14.6} \\
\bottomrule
\end{tabular}
\caption{SHIP with ProbeSelect on SD3.5-Large. All rows use GenEval2 and the same $400/400$ split. Full-denoise equivalents normalize denoising cost. Empirical UCB values describe calibration feasibility rather than the fixed-grid certificate.}
\label{tab:probeselect}
\end{table*}

\begin{table*}[t]
\centering
\small
\setlength{\tabcolsep}{5.0pt}
\renewcommand{\arraystretch}{1.08}
\begin{tabular}{@{}llrrrrrr@{}}
\toprule
Editing policy & Calibration unit & Cov.$\uparrow$ & Edit loss$\downarrow$ & Cal. UCB$\downarrow$ & NFE$\downarrow$ & VLM/p$\downarrow$ & Sec./edit$\downarrow$ \\
\midrule
Best-of-$N$ editing & None & 1.000 & 0.281 & N/A & 896 & 192 & 208.6 \\
ADE-CoT & Native stopping & 1.000 & 0.274 & N/A & 441 & 94 & 102.7 \\
ADE-CoT & Candidate & 0.564 & 0.243 & 0.408 & 441 & 94 & 102.7 \\
\textbf{ADE-CoT + SHIP} & Policy & 0.472 & \textbf{0.171} & \textbf{0.298} & 441 & 94 & 102.7 \\
\bottomrule
\end{tabular}
\caption{Cross-task policy calibration for image editing. FLUX.1-Kontext-dev and $N{=}32$ follow the ADE-CoT protocol. SHIP changes the release rule without changing the native ADE-CoT trajectory cost.}
\label{tab:editing}
\end{table*}

The adaptive results distinguish search quality from release calibration.
Table~\ref{tab:matched-risk} shows that the policies occupy different points
on the same matched-risk frontier. Tables~\ref{tab:probeselect}
and~\ref{tab:editing} then test early trajectory selection and iterative image
editing. SHIP changes which completed policy outcomes are released. It does
not claim the native search contributions or their internal compute savings.

\section{Robustness across Models, Data, Judges, and Splits}
\label{sec:supp-robustness}

\begin{figure*}[t]
\centering
\includegraphics[width=0.78\textwidth]{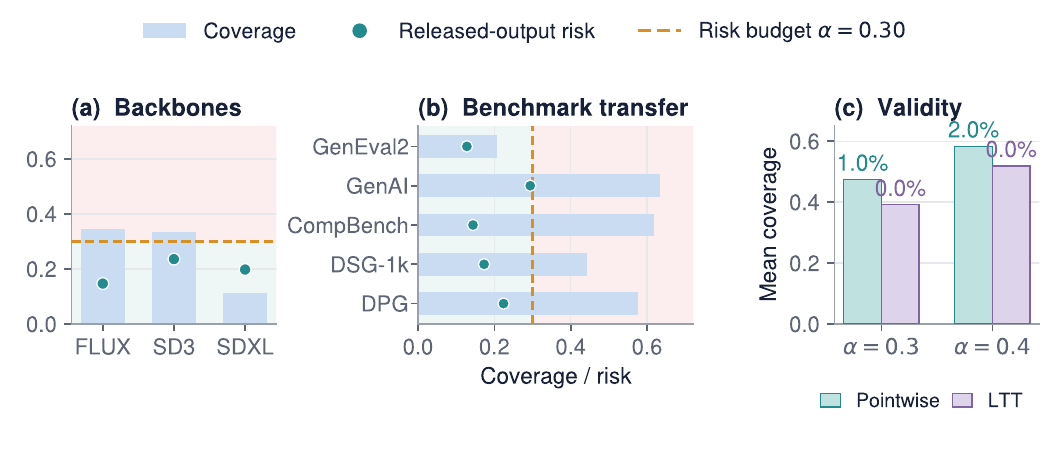}
\caption{Robustness summary. \emph{a} Held-out coverage and risk across three T2I backbones. \emph{b} Fixed-threshold transfer from GenEval2 to four FLUX prompt sets. \emph{c} Mean coverage and target-crossing rates over $200$ cached-stream splits.}
\label{fig:robustness}
\end{figure*}

\paragraph{Prompt-distribution transfer.}
Table~\ref{tab:benchmark-transfer} expands the middle panel of
Figure~\ref{fig:robustness}. The release threshold $\tau{=}0.322$ is fit on
GenEval2 seed-$0$ calibration images and transferred unchanged to the seed-$0$
FLUX images for each prompt set. Transfer remains below the nominal budget in
these four cases, but the
coverage and risk shifts show why the result should not be interpreted as a
distribution-shift guarantee.

\begin{table}[t]
\centering
\small
\setlength{\tabcolsep}{3.4pt}
\renewcommand{\arraystretch}{1.06}
\begin{tabular}{@{}lrr@{}}
\toprule
Prompt set & Cov.$\uparrow$ & Risk$\downarrow$ \\
\midrule
GenEval2 test & 0.207 & 0.128 \\
GenAI-Bench & 0.633 & 0.294 \\
T2I-CompBench & 0.619 & 0.144 \\
DSG-1k & 0.443 & 0.173 \\
DPG-Bench & 0.576 & 0.224 \\
\bottomrule
\end{tabular}
\caption{Single-candidate FLUX prompt-distribution transfer with a fixed GenEval2-calibrated threshold. No transfer set is used to select the displayed threshold.}
\label{tab:benchmark-transfer}
\end{table}

\paragraph{Alternative target judge.}
Table~\ref{tab:target-judge} replaces the Qwen3 target with InternVL3 while
keeping the Qwen2.5 selector fixed. The selector correlates strongly with both
targets, and policy-level calibration preserves the ordering of accept or
reject, best-of-$4$, and sequential policies. Absolute risks change, which
reinforces the interpretation that the target judge defines the measured
loss.

\begin{table}[t]
\centering
\small
\setlength{\tabcolsep}{3.2pt}
\renewcommand{\arraystretch}{1.06}
\begin{tabular}{@{}lrrrr@{}}
\toprule
Target and policy & $\rho$ & Cov.$\uparrow$ & Risk$\downarrow$ & Gen/p$\downarrow$ \\
\midrule
Qwen3, accept or reject & 0.922 & 0.210 & 0.127 & 1.00 \\
Qwen3, best-of-$4$ + SHIP & 0.922 & 0.403 & 0.192 & 4.00 \\
Qwen3, sequential SHIP & 0.922 & 0.347 & 0.147 & 3.18 \\
\midrule
InternVL3, accept or reject & 0.897 & 0.290 & 0.166 & 1.00 \\
InternVL3, best-of-$4$ + SHIP & 0.897 & 0.573 & 0.232 & 4.00 \\
InternVL3, sequential SHIP & 0.897 & 0.438 & 0.172 & 2.95 \\
\bottomrule
\end{tabular}
\caption{Alternative-target robustness at $N{=}4$. The symbol $\rho$ is the Spearman correlation between Qwen2.5 selector risk and the stated target loss over $12800$ candidates. Each target is calibrated separately.}
\label{tab:target-judge}
\end{table}

\paragraph{Pointwise operating points and fixed-grid certification.}
Table~\ref{tab:certificate} separates the practical pointwise scan used in the
main result tables from the prespecified-grid procedure in Proposition~2.
Both rows use one prompt trajectory as one observation and average over $200$
cached-stream splits at $N{=}16$. The fixed-grid row has no empirical target
crossing and is the only row in this comparison with simultaneous
finite-sample validity. The pointwise row attains higher mean coverage but is
reported only as an empirical operating rule.

\begin{table}[t]
\centering
\small
\setlength{\tabcolsep}{3.2pt}
\renewcommand{\arraystretch}{1.06}
\begin{tabular}{@{}lrrr@{}}
\toprule
Threshold rule & Mean cov. & Mean risk & Crossings \\
\midrule
Pointwise Hoeffding & 0.474 & 0.226 & $2/200$ \\
Fixed-grid SHIP & 0.391 & 0.148 & $0/200$ \\
\bottomrule
\end{tabular}
\caption{Sequential $N{=}16$ split stability at $\alpha{=}0.30$. The fixed-grid rule uses $201$ prespecified thresholds with Bonferroni allocation. Crossings count held-out test risk point estimates above $\alpha$ across cached-stream splits.}
\label{tab:certificate}
\end{table}

\begin{table*}[t]
\centering
\small
\setlength{\tabcolsep}{3.2pt}
\renewcommand{\arraystretch}{1.08}
\begin{tabular}{@{}lllrrrl@{}}
\toprule
Baseline & Native benchmark & Official metric & Paper & Reproduced & Rel. gap & Fidelity note \\
\midrule
ProbeSelect & SD3.5-Large & ImageReward, base to selected & 1.14 to 1.83 & 1.16 to 1.74 & $-4.9\%$ & Probe head retrained \\
ProbeSelect & SD3.5-Large & HPSv2.1, base to selected & 30.29 to 31.81 & 30.34 to 31.55 & $-0.8\%$ & Same probe head \\
ProbeSelect & SD3.5-Large & Analytic cost ratio & 0.36 & 0.36 & $0.0\%$ & Exact \\
Flash-BoN & GenAI-Bench & Normalized AUC & 0.68 & 0.65 & $-4.4\%$ & Hardware-dependent time \\
RAISE & GenEval & Overall score & 0.94 & 0.92 & $-2.1\%$ & Rewriter sampling \\
RAISE & GenEval & Samples per prompt & 18.6 & 19.4 & $+4.3\%$ & Verifier variation \\
RAISE & GenEval & VLM calls per prompt & 7.3 & 7.8 & $+6.8\%$ & Verifier variation \\
RAISE & DrawBench & VQAScore & 0.885 & 0.877 & $-0.9\%$ & Exact protocol \\
RAISE & DrawBench & ImageReward & 1.15 & 1.11 & $-3.5\%$ & Exact protocol \\
RAISE & DrawBench & HPS-v2 & 0.305 & 0.302 & $-1.0\%$ & Exact protocol \\
ADE-CoT & GEdit-Bench & $G_O$, base to ADE-CoT & 6.641 to 6.695 & 6.52 to 6.58 & $-1.7\%$ & Judge-version drift \\
ADE-CoT & Kontext & NFE & 418 & 441 & $+5.5\%$ & Difficulty estimator \\
\bottomrule
\end{tabular}
\caption{Reproduction fidelity for adaptive baselines. Official metrics are evaluated on each method's native benchmark before the common SHIP protocol. Small residual gaps arise from the stated implementation or hardware difference.}
\label{tab:fidelity}
\end{table*}

\section{Qualitative Evidence from Real Generations}
\label{sec:supp-qualitative}

\begin{figure*}[t]
\centering
\includegraphics[width=0.92\textwidth]{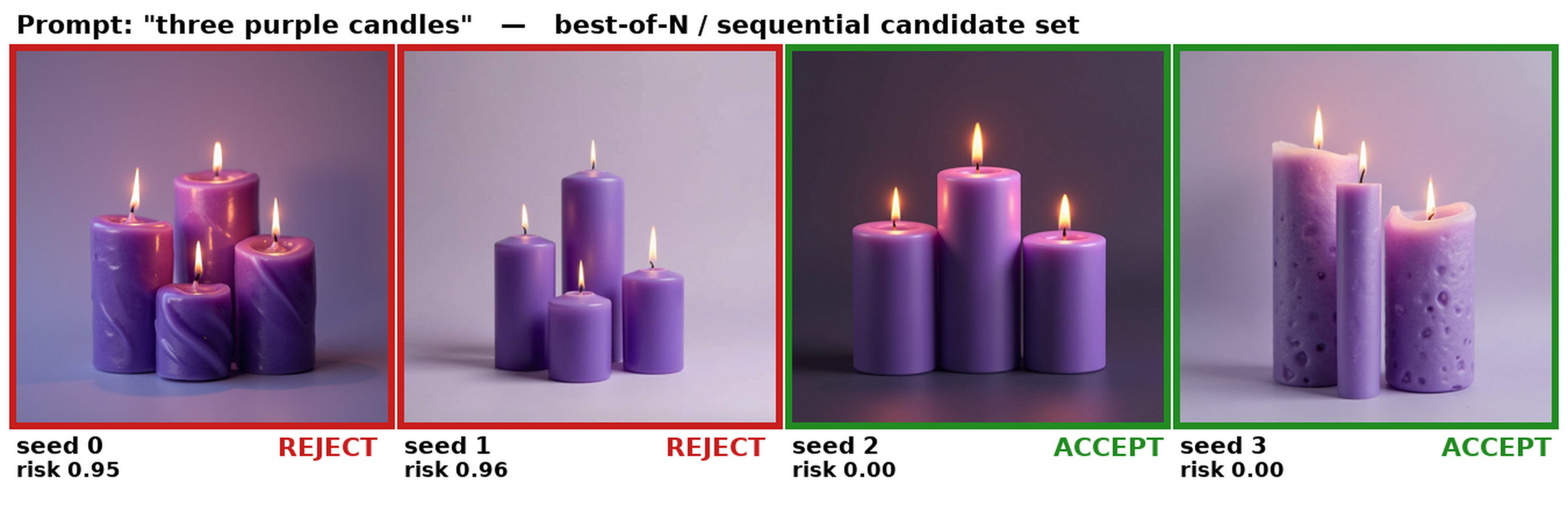}\\
\includegraphics[width=0.98\textwidth]{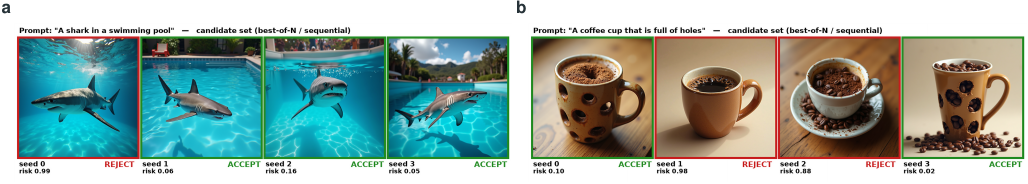}
\caption{Qualitative gallery of real FLUX candidate streams with Qwen2.5
selector risk. \emph{Top}: for ``three purple candles,'' four-candle renders
are rejected and three-candle renders are accepted. \emph{Bottom left}: an
open-ocean shark is rejected for a swimming-pool prompt. \emph{Bottom right}:
ordinary or bean-filled cups are rejected for a cup-full-of-holes prompt. The
held-out target agrees on every displayed decision. Sequential stopping
releases the first passing candidate in each row.}
\label{fig:gallery}
\end{figure*}

Figure~\ref{fig:gallery} makes the release rule visible at the image level.
The candidate streams contain both clear failures and faithful generations
for counting, contextual, and structural requirements. The figure is
diagnostic rather than a selected success rate. Quantitative evidence is
always computed over the complete held-out prompt set.

\section{Fixed Search, Evaluators, and Failure Diagnostics}
\label{sec:supp-fixed}

\paragraph{Risk and coverage across target budgets.}
Table~\ref{tab:alpha-sweep} reports every operating point shown on the
risk-coverage frontier in the main paper. The threshold is calibrated
separately for each target budget. Sequential stopping and best-of-$4$
rejection use the same candidate bank and target judge. Sequential stopping
reduces generation and QA cost because accepted prompts may terminate before
the fourth attempt.

\begin{table*}[t]
\centering
\small
\setlength{\tabcolsep}{4.3pt}
\renewcommand{\arraystretch}{1.06}
\begin{tabular}{@{}clrrrr@{}}
\toprule
$\alpha$ & Policy & Cov.$\uparrow$ & Risk$\downarrow$ & Gen/p$\downarrow$ & QA/p$\downarrow$ \\
\midrule
0.15 & Sequential SHIP & 0.185 & 0.033 & 3.62 & 28.0 \\
0.20 & Sequential SHIP & 0.253 & 0.068 & 3.43 & 26.8 \\
0.25 & Sequential SHIP & 0.320 & 0.119 & 3.25 & 25.7 \\
0.30 & Sequential SHIP & 0.347 & 0.147 & 3.18 & 25.1 \\
0.35 & Sequential SHIP & 0.403 & 0.226 & 3.05 & 24.2 \\
0.40 & Sequential SHIP & 0.438 & 0.295 & 2.95 & 23.4 \\
0.50 & Sequential SHIP & 0.557 & 0.425 & 2.68 & 21.2 \\
\midrule
0.15 & Best-of-$4$ + SHIP & 0.185 & 0.032 & 4.00 & 30.1 \\
0.20 & Best-of-$4$ + SHIP & 0.273 & 0.063 & 4.00 & 30.1 \\
0.25 & Best-of-$4$ + SHIP & 0.330 & 0.102 & 4.00 & 30.1 \\
0.30 & Best-of-$4$ + SHIP & 0.403 & 0.192 & 4.00 & 30.1 \\
0.35 & Best-of-$4$ + SHIP & 0.485 & 0.280 & 4.00 & 30.1 \\
0.40 & Best-of-$4$ + SHIP & 0.552 & 0.339 & 4.00 & 30.1 \\
0.50 & Best-of-$4$ + SHIP & 0.680 & 0.442 & 4.00 & 30.1 \\
\bottomrule
\end{tabular}
\caption{Complete empirical risk, coverage, and cost frontier for FLUX at $N{=}4$. These are pointwise operating points. The requested risk budget changes by row and is not a shared horizontal constraint across the table.}
\label{tab:alpha-sweep}
\end{table*}

\paragraph{Budget sweep.}
Table~\ref{tab:budget} varies the maximum number of candidates at the default
risk budget. Coverage rises with $N$, while compute grows more slowly than the
maximum budget because the sequential policy stops after its first pass.

\paragraph{Cross-Backbone Operating Points.} Table~\ref{tab:backbone} gives the per-backbone numbers summarized in the main text. Held-out risk is below budget for all three generators, while achievable coverage varies substantially.

\paragraph{Where Failures Concentrate.} Table~\ref{tab:skill} breaks accepted-output quality down by atom type, showing that counting, spatial, and verb atoms are the bottleneck while object existence is nearly solved.

\paragraph{Adaptive-policy generalization.}
Tables~\ref{tab:matched-risk}, \ref{tab:probeselect}, and~\ref{tab:editing}
provide the complete operating points summarized in the main paper.
Table~\ref{tab:fidelity} separately checks each reproduction on its native
benchmark before applying the shared SHIP evaluation protocol.

\begin{table}[H]
\centering
\small
\renewcommand{\arraystretch}{1.05}
\begin{tabular}{@{}ccccc@{}}
\toprule
$N$ & Cov.$\uparrow$ & Risk$\downarrow$ & Gen/p$\downarrow$ & QA/p$\downarrow$ \\
\midrule
1 & 0.207 & 0.128 & 1.00 & 7.5 \\
2 & 0.297 & 0.151 & 1.78 & 13.7 \\
4 & 0.347 & 0.147 & 3.18 & 25.1 \\
8 & 0.390 & 0.168 & 5.80 & 46.8 \\
16 & 0.438 & 0.162 & 10.68 & 87.6 \\
\bottomrule
\end{tabular}
\caption{Budget sweep for the empirical sequential operating point on FLUX. The risk budget is $\alpha{=}0.30$. Each budget is calibrated on the complete first-pass policy.}
\label{tab:budget}
\end{table}

\begin{table}[H]
\centering
\small
\renewcommand{\arraystretch}{1.05}
\begin{tabular}{@{}lccc@{}}
\toprule
Backbone & Cov.$\uparrow$ & Risk$\downarrow$ & Gen/p$\downarrow$ \\
\midrule
FLUX.1-dev & 0.347 & 0.147 & 3.18 \\
SD3 Medium & 0.335 & 0.236 & 3.29 \\
SDXL & 0.113 & 0.198 & 3.76 \\
\bottomrule
\end{tabular}
\caption{Cross-backbone empirical operating points for the sequential policy at $\alpha{=}0.30$. Held-out risk is below the budget on all three backbones, while coverage varies substantially.}
\label{tab:backbone}
\end{table}

\begin{table}[H]
\centering
\small
\renewcommand{\arraystretch}{1.05}
\begin{tabular}{@{}lcc@{}}
\toprule
Atom type & Target score$\uparrow$ & Failure rate$\downarrow$ \\
\midrule
Object existence & 0.956 & 4.3\% \\
Counting & 0.706 & 29.6\% \\
Spatial position & 0.523 & 47.6\% \\
Verb / action & 0.264 & 70.7\% \\
\bottomrule
\end{tabular}
\caption{Accepted-output quality by atom type under best-of-$8$ with calibrated rejection. Counting, spatial, and verb atoms dominate residual failures.}
\label{tab:skill}
\end{table}

\section{Human Released-Output Audit}
\label{sec:supp-human}

The audit evaluates both $N{=}16$ sequential policies on the same $400$ test
prompts. Their released sets contain $267$ distinct images and share $137$
outputs. Blinded views omit policies, thresholds, and machine scores. Two
annotators label each atom as yes, no, or uncertain, followed by adjudication
of disagreements. Shared outputs are labeled once and reused.

Atomic loss is the fraction of adjudicated atoms not labeled yes.
Any-failure marks an image with at least one no or uncertain atom. Intervals
resample the full prompt universe while preserving release and abstention.
Because machine loss selects the thresholds, Table~\ref{tab:human}
corroborates the result but does not certify human risk.

\begin{table}[H]
\centering
\scriptsize
\setlength{\tabcolsep}{1.7pt}
\renewcommand{\arraystretch}{1.02}
\begin{tabular}{@{}lrrr@{}}
\toprule
Metric & Pooled & SHIP & Difference \\
\midrule
Released outputs & 229 & \textbf{175} & N/A \\
Coverage & 0.573 & \textbf{0.438} & $-0.135$ \\
Atomic loss & 0.112 & \textbf{0.076} & $-0.036$ \\
$95\%$ CI & [0.090, 0.134] & \textbf{[0.057, 0.095]} & [$-0.057$, $-0.015$] \\
Any-failure & 0.380 & \textbf{0.280} & $-0.100$ \\
$95\%$ CI & [0.315, 0.445] & \textbf{[0.212, 0.348]} & [$-0.147$, $-0.053$] \\
Uncertain atoms & 0.031 & \textbf{0.021} & $-0.010$ \\
\bottomrule
\end{tabular}
\caption{Human audit of released policy outputs on the same $400$ prompts. The difference is SHIP minus pooled-candidate sequential release. Atomic loss uses $1550$ pooled-policy atoms and $1102$ SHIP-policy atoms. Intervals use prompt bootstrap.}
\label{tab:human}
\end{table}

\begin{table}[H]
\centering
\small
\renewcommand{\arraystretch}{1.05}
\begin{tabular}{@{}lr@{}}
\toprule
Statistic & Value \\
\midrule
Raw agreement & 0.880 \\
Cohen's $\kappa$ & 0.460 \\
PABAK & 0.760 \\
Gwet's AC1 & 0.865 \\
Positive specific agreement & 0.610 \\
Adjudication rate & 0.120 \\
\bottomrule
\end{tabular}
\caption{Human annotation reliability. Two primary labels were collected for each of $1{,}806$ unique atom decisions, producing $3{,}612$ primary annotations.}
\label{tab:human-agreement}
\end{table}

\section{Evaluator Ablations and Modular Cascades}
\label{sec:supp-evaluator}

\begin{table}[H]
\centering
\setlength{\tabcolsep}{2.7pt}
\renewcommand{\arraystretch}{1.04}
\small
\begin{tabular}{@{}lrrrr@{}}
\toprule
Selector & $\rho\uparrow$ & AUROC$\uparrow$ & Cov.$\uparrow$ & Risk$\downarrow$ \\
\midrule
\multicolumn{5}{@{}l}{\emph{Fine-grained atom-wise risk}} \\
\textbf{Soft-TIFA-GM} & \textbf{0.928} & 0.975 & 0.403 & \textbf{0.192} \\
Soft-TIFA-AM & 0.860 & 0.950 & 0.400 & 0.229 \\
Max-atom risk & 0.814 & 0.976 & 0.390 & 0.195 \\
Learned linear aggregation & 0.884 & \textbf{0.978} & \textbf{0.420} & 0.222 \\
\addlinespace[2pt]
\multicolumn{5}{@{}l}{\emph{Holistic similarity or preference}} \\
VQAScore & 0.741 & 0.911 & 0.000 & N/A \\
ImageReward & 0.420 & 0.685 & 0.000 & N/A \\
HPSv2 & 0.351 & 0.685 & 0.000 & N/A \\
PickScore & 0.346 & 0.659 & 0.000 & N/A \\
CLIPScore & 0.135 & 0.474 & 0.000 & N/A \\
\midrule
Target-loss oracle & 1.000 & 1.000 & 0.510 & 0.229 \\
\bottomrule
\end{tabular}
\caption{Selector ablation on the held-out target. The symbol $\rho$ denotes Spearman correlation. Coverage and risk are measured for best-of-$4$ policy-level rejection at $\alpha{=}0.30$. Only fine-grained atom-wise selectors yield a nonempty release set under the empirical feasibility criterion.}
\label{tab:eval}
\end{table}

\paragraph{Cheap-to-expensive selector cascades.}
The release gate need not score every candidate with the fine-grained
evaluator. Table~\ref{tab:cascade} first ranks four candidates with a cheaper
holistic score, then sends only the top one or two candidates to the
Soft-TIFA-GM gate. Every row uses the same target judge and policy-level
release calibration. Sending two candidates improves coverage while using
half of the $30.1$ atom-level QA calls required by scoring all four candidates.

\begin{table}[H]
\centering
\small
\setlength{\tabcolsep}{3.0pt}
\renewcommand{\arraystretch}{1.06}
\begin{tabular}{@{}lrrrr@{}}
\toprule
Prefilter & Top $k$ & Cov.$\uparrow$ & Risk$\downarrow$ & QA/p$\downarrow$ \\
\midrule
VQAScore & 1 & 0.340 & 0.176 & 7.5 \\
VQAScore & 2 & 0.388 & 0.187 & 15.0 \\
CLIPScore & 1 & 0.225 & 0.099 & 7.5 \\
CLIPScore & 2 & 0.320 & 0.168 & 15.0 \\
HPSv2 & 1 & 0.260 & 0.166 & 7.5 \\
HPSv2 & 2 & 0.352 & 0.195 & 15.0 \\
PickScore & 1 & 0.255 & 0.137 & 7.5 \\
PickScore & 2 & 0.320 & 0.165 & 15.0 \\
ImageReward & 1 & 0.250 & 0.119 & 7.5 \\
ImageReward & 2 & 0.340 & 0.158 & 15.0 \\
\bottomrule
\end{tabular}
\caption{Cheap-to-expensive cascade at $N{=}4$ and $\alpha{=}0.30$. Every prefilter uses four cheap calls. Soft-TIFA-GM still makes the calibrated release decision.}
\label{tab:cascade}
\end{table}

\section{Controlled Study and Deployment Boundaries}
\label{sec:supp-boundaries}

\paragraph{Synthetic controlled study and a lower bound.}
To isolate selection from any particular T2I verifier, Figure~\ref{fig:synthetic}a
uses a two-component prompt mixture. A hard prompt occurs with probability
$0.5$ and has candidate failure probability $0.9$, while an easy prompt has
failure probability $0.05$. The selector score is
$s=\ell+\varepsilon$, where $\varepsilon\sim\mathcal{N}(0,0.7^2)$. For every
candidate budget, we choose $\tau$ so
$\mathbb{E}[\ell\mid s\le\tau]=\alpha$ and then deploy best-of-$N$ or sequential
selection. Candidate calibration remains at $0.30$, yet released risk can
exceed the target. Policy calibration stays at or below the target.

Figure~\ref{fig:synthetic}b studies the limiting construction used below. A
prompt is hard with probability $\rho$. Every hard-prompt candidate has loss
$1$, every easy-prompt candidate has loss $0$, and
$\varepsilon\sim\mathcal{N}(0,1.5^2)$. With $N{=}256$, released risk reaches
$0.64$ at $\rho{=}0.7$ while candidate risk remains $0.30$.

\medskip\noindent\textbf{Proposition 3} (Lower bound on the post-selection gap)\textbf{.}
\emph{In the deterministic-loss mixture of Figure~\ref{fig:synthetic}b, fix any candidate-level threshold $\tau$ with $\mathbb{E}[\ell\mid s\le\tau]=\alpha$. As $N\to\infty$, the best-of-$N$ released-output risk under $\tau$ converges to the hard-prompt share $\rho$ among accepted prompts, while candidate-level risk remains $\alpha$. Hence the gap $R_\pi(\tau)-R_{\mathrm{img}}(\tau)$ can be driven to $1-\alpha$ by increasing $\rho$.}
\noindent\emph{Proof sketch.} As $N\to\infty$, $\min_{k\in[N]}s(Y_k,p)\to-\infty$, so acceptance approaches one. Hard prompts then release loss $1$ and easy prompts release loss $0$, giving $R_\pi(\tau)\to\rho$, while the fixed candidate risk remains $\alpha$. Taking $\rho\to1$ yields a gap approaching $1-\alpha$. $\square$ Thus candidate calibration cannot in general be patched into a policy guarantee.

\begin{figure}[t]
\centering
\includegraphics[width=0.98\columnwidth]{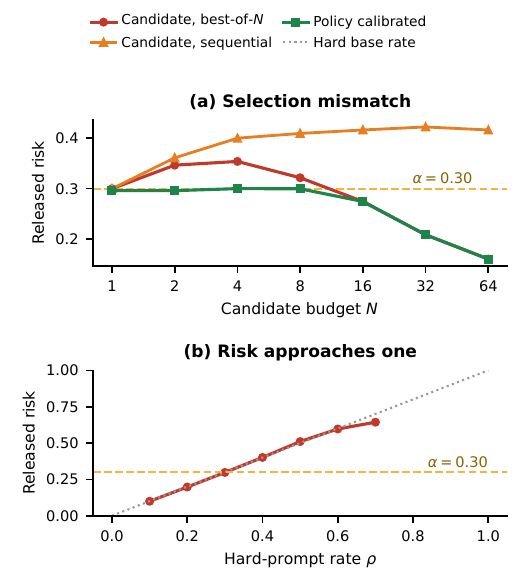}
\caption{Controlled study. \emph{a} Candidate calibration meets $\alpha$, but released risk after selection can exceed it. \emph{b} Best-of-$N$ risk tracks the hard-prompt rate $\rho$ and approaches the lower bound in Proposition~3.}
\label{fig:synthetic}
\end{figure}

\paragraph{Group-Conditional Risk Control.}
A global threshold controls average released risk, not subgroup risk. Table~\ref{tab:group} groups prompts by atomicity. The $9$ to $10$ atom group has risk $0.398$ at $0.05$ coverage from five releases. Group-SHIP calibrates each group separately and abstains where no threshold is feasible, so it remains a diagnostic rather than a service guarantee.

\begin{table}[t]
\centering\small
\renewcommand{\arraystretch}{1.05}
\begin{tabular}{@{}lcccc@{}}
\toprule
& \multicolumn{2}{c}{Global $\tau$} & \multicolumn{2}{c}{Group-SHIP} \\
Atoms & Cov.$\uparrow$ & Risk$\downarrow$ & Cov.$\uparrow$ & Risk$\downarrow$ \\
\midrule
3 to 4 & 0.66 & 0.097 & 0.73 & 0.150 \\
5 to 6 & 0.50 & 0.147 & 0.40 & 0.112 \\
7 to 8 & 0.18 & 0.263 & 0.00 & N/A \\
9 to 10 & 0.05 & \textbf{0.398} & 0.00 & N/A \\
\bottomrule
\end{tabular}
\caption{Group-conditional sequential risk at $\alpha{=}0.30$. The hardest group has risk $0.398$ from five releases. Group-SHIP abstains where no threshold is feasible.}
\label{tab:group}
\end{table}

\paragraph{Per-Deployment Recalibration.}
The GenAI-Bench transfer is the tightest case. On SD3, the transferred risk point estimate reaches $0.303>\alpha$. Table~\ref{tab:recalib} shows that a small deployment-side calibration set returns the average held-out risk below $\alpha$, with coverage recovering as $n_{\mathrm{cal}}$ grows.

\begin{table}[t]
\centering\small
\renewcommand{\arraystretch}{1.05}
\begin{tabular}{@{}lccc@{}}
\toprule
$n_{\mathrm{cal}}$ & 50 & 100 & 200 \\
\midrule
Test risk$\downarrow$ & 0.053 & 0.139 & 0.197 \\
Coverage$\uparrow$ & 0.173 & 0.419 & 0.544 \\
\bottomrule
\end{tabular}
\caption{GenAI-Bench recalibration with SD3 Medium over $200$ splits.}
\label{tab:recalib}
\end{table}

\paragraph{Sensitivity and Finite-Sample Certification.}
The threshold is stable across tested resolutions. Coverage is $0.36$ at $|\mathcal{T}|{=}25$ and $0.39$ at full resolution. Calibration sets of $200$ to $400$ prompts yield coverage from $0.36$ to $0.39$, with cached-stream crossing rates below $\delta$. Policy recalibration lowers released risk over pooled calibration by $0.087$ for FLUX, $0.096$ for SD3, and $0.166$ for SDXL at $N{=}4$. One fixed candidate per prompt gives $\tau{=}0.322$ and, under sequential $N{=}16$ reuse, risk $0.196$ with interval $[0.156,0.239]$ at coverage $0.473$. Thus the empirical crossing depends on the candidate design, while the population non-implication remains. Empirical-Bernstein did not improve coverage under grid correction. Anytime-valid betting increased coverage but did not reliably remain below $\delta$. Sharper selection-aware concentration remains future work.

\FloatBarrier
\fi
\end{document}